\documentclass[journal]{IEEEtran}

\usepackage[ruled,vlined,linesnumbered]{algorithm2e}
\usepackage{graphicx,subfigure}
\usepackage{booktabs}
\usepackage{amsmath}
\usepackage{multirow}
\usepackage[numbers,sort&compress]{natbib}
\usepackage{array}
\usepackage{epsfig}
\usepackage{epstopdf}
\newcolumntype{L}[1]{>{\raggedright\let\newline\\\arraybackslash\hspace{0pt}}m{#1}}
\newcolumntype{C}[1]{>{\centering\let\newline\\\arraybackslash\hspace{0pt}}m{#1}}
\newcolumntype{R}[1]{>{\raggedleft\let\newline\\\arraybackslash\hspace{0pt}}m{#1}}

\usepackage{amsthm}
\usepackage{fixmath}
\usepackage{hyperref}
\usepackage{xcolor}
\definecolor{hyperlink}{rgb}{0,0,93.3}

\DeclareMathOperator*{\argmin}{arg\,min}

\usepackage{mathtools} 
\usepackage{siunitx}

\usepackage{amsmath,amssymb}

\usepackage{tikz}
\usetikzlibrary{matrix,shapes,arrows,positioning,chains}

\usepackage{url}

\begin{document}

\title{Jack and Masters of All Trades: One-Pass Learning of a Set of Model Sets from Foundation AI Models}  

 \author{Han~Xiang~Choong,~Yew-Soon~Ong,~Abhishek~Gupta~and~Ray~Lim
\thanks{Han Xiang Choong is with the School of Computer Science and Engineering, Nanyang Technological University, Singapore
 (e-mail: hanxiang001@e.ntu.edu.sg).}
\thanks{Yew-Soon Ong is with the School of Computer Science and Engineering, Nanyang Technological University, and also with the Agency for Science, Technology and Research, Singapore (e-mail: ASYSONG@ntu.edu.sg).} 
 \thanks{Abhishek Gupta is with the Singapore Institute of Manufacturing Technology, of the Agency for Science, Technology and Research, Singapore
 (e-mail: ABHISHEK\_GUPTA@simtech.a-star.edu.sg).}
 \thanks{Ray Lim is with the School of Computer Science and Engineering, Nanyang Technological University, Singapore
 (e-mail: ray.lim@ntu.edu.sg).}
 }
\maketitle

\begin{abstract}
For deep learning, size is power. Massive neural nets trained on broad data for a spectrum of tasks are at the forefront of artificial intelligence. These foundation models or ’Jacks of All Trades’ (JATs), when fine-tuned for downstream tasks, are gaining importance in driving deep learning advancements. However, environments with tight resource constraints, changing objectives and intentions, or varied task requirements, could limit the real-world utility of a singular JAT. Hence, in tandem with current trends towards building increasingly large JATs, this paper conducts an initial exploration into concepts underlying the creation of a diverse set of compact machine learning model sets. Composed of many smaller and specialized models, we formulate the \emph{Set of Sets} to simultaneously fulfil many task settings and environmental conditions. A means to arrive at such a set tractably in one pass of a \emph{neuroevolutionary multitasking} algorithm is presented for the first time, bringing us closer to models that are collectively 'Masters of All Trades'. 
\end{abstract}

\IEEEpeerreviewmaketitle
\section{Introduction} 

Buoyed by bountiful computing power and data, \textbf{Deep Neural Networks (DNNs)} currently enjoy primacy in \textbf{Machine Learning (ML)}. From early DNNs succeeding at highly specific and narrow tasks [1], the field has progressed gradually towards increasingly general models. Consequently, the size and complexity of DNNs has skyrocketed [2]. Consider the original LeNet (65K parameters) [3], as well as AlphaGo (4.6M parameters) [4]. In performing numeral recognition and in playing Go, these models are specialized for a single task, and may be called \textbf{Masters of One Trade (MOTs)}. In contrast, ViT-G/14 [5] (1.8B parameters), and GPT-3 (175B parameters) [6], are capable of a virtually unlimited range of tasks within \textit{Computer Vision (CV)} and \textit{Natural Language Processing (NLP)}. These latter examples have come to be known as foundation models [7], approaching our notion of \textbf{Jacks of All Trades (JATs)} in this paper. Being trained on broad data at scale, these models play a key supporting role in ML development through downstream transfer learning [8]. 

The archetypal JAT (Fig. 1) is massive and highly expressive [9], allowing the learning of multiple tasks simultaneously [10]. Such training exploits similarities between tasks, promotes the acquisition of generalizable knowledge, and thus improves performance [11].  However, this multifaceted learning sometimes results in a trading-off of errors between tasks, as a natural consequence of capacity limits in practice. This limitation could hinder performance of individual tasks [12]. Depending on the complexity, size, and number of tasks to be learned, JATs with finite information encoding capacity often cannot afford to allocate each task a specialized internal substructure. Conflicts between task-specific training signals could therefore occur, causing JATs to be weaker at performing individual tasks compared to specialists. 

In addition, despite the flexibility conferred by greater generality, JATs are often too large to be deployed in many situations. Resource-constrained computing requirements are growing in prevalence [13--15] owing to the surge in popularity of deep learning across industries and fields. As a result, the problem of inaccessibility---i.e., the lack of resources to utilize such state-of-the-art tools---is recognized as being of critical importance to the majority of the community [16, 17]. Nevertheless, the pursuit of greater generality continues unabated, with the size and complexity of models ever-increasing.

\begin{figure*}[!htbp]
  \includegraphics[width=\linewidth]{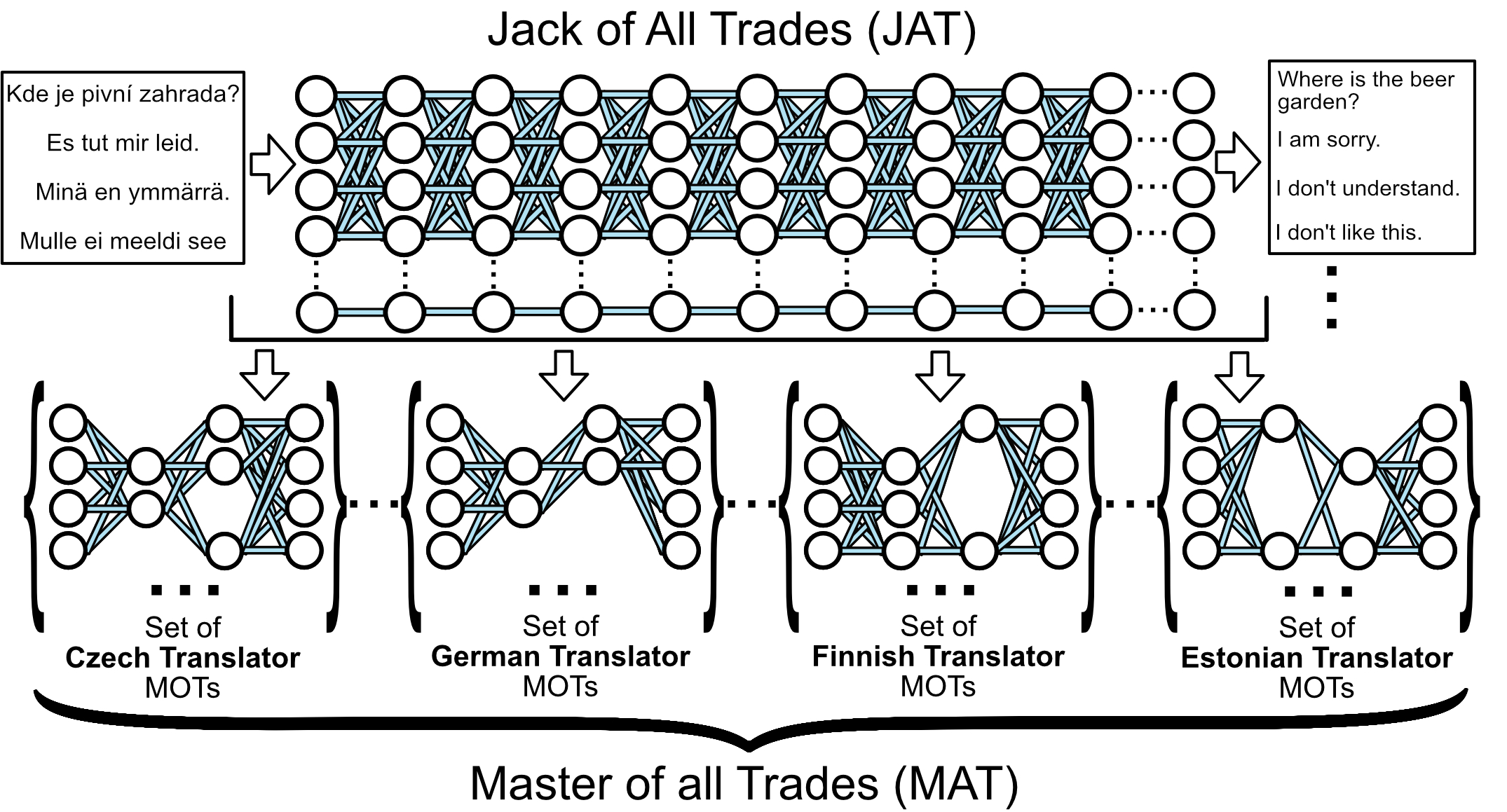}
  \caption{An illustrative application in multilingual translation. The JAT is a singular model that performs many translation tasks. Each compact MOT specializes for a single translation task. By originating from the JAT, MOTs shed conflicting information and gain generalizable knowledge. In this paper, we investigate whether the collective of MOTs could surpass the JAT in performance and computational efficiency, gradually approximating a \textbf{Master of All Trades (MAT)}.}
  \label{fig:multi-task}
\end{figure*}

The culmination of this pursuit is envisioned to be a model that exhibits broad, human-level intelligence (or beyond) across a spectrum of tasks, often claimed to constitute AI's Holy Grail [18]. The existence of at least two associated pathways that could lead to this vision is conjectured. The first path is clearly marked by current trends towards the building of titanic DNNs, brute-forcing the mastery of many tasks using sheer size. These effectively unconstrained models shall be equipped to possess a very large number of task-specialized components (when needed), would not suffer from any capacity-related conflicts, and could therefore be classified as \textbf{Masters of All Trades (MATs)}. However, this direction is encumbered by both extreme computational expense that's inaccessible to most, and diminishing returns [19]. What's more, unfettered generality is often not required in practice. To illustrate, a classical opera may be in need of a violinist, but is far less likely to specifically need a violinist who also plays the guitar and the trombone. 

This paper presents a first exploration of a complementary pathway. The archetypal MOT is a task specialist; compact, efficient, and focused on a given task. The less complex nature of its structure and training suggests reduced internal conflict relative to JATs. This theoretically allows MOTs to achieve stronger performance in their specific domains. \emph{If an MOT can sufficiently surpass the performance of a JAT on relatively narrow tasks, then a collective of such MOTs, over many tasks, starts to functionally approach an MAT}. A collective is modular, can easily grow and shrink, whilst adapting to a priori unknown objectives, intentions and constraints of its human end-users. Given current materials, methods, and technologies, such a collective could be produced economically, with sizeable resource efficiency advantages in the long-term.

Importantly, MOTs, JATs, and MATs are conceived as being interdependent rather than separate. The creation of many task-specialized DNNs is proposed by taking into consideration a number of computational factors. One of these is the considerable energy cost incurred by a single development cycle in deep learning [20]. Another factor is the potential to harness valuable information contained within the parameters of pre-trained JATs. When trained from scratch for narrow tasks, an MOT does not benefit from a broad base of fundamental and generalizable knowledge as a JAT would. On the other hand, by learning MOTs from well-trained JATs, i.e. training a large model and then cutting it down small [21], we can alleviate these concerns. An abstraction of this concept is depicted in Fig. 1, where a massive multilingual translation model is compressed into sets of small but specialized monolingual translators. The collective of many such MOTs, each suited for a different environment and task, constitutes what we refer to as the \textbf{set of ML model sets (or Set of Sets)}.

The creation and deployment of a Set of Sets promises attractive benefits, especially in resource-constrained environments. It is then necessary to devise methods that could generate such a set in a tractable manner. Going forward, we thus aim to provide a novel means to arrive at optimized sets of ML models (i.e., Sets of Masters, MOTs)---that simultaneously fulfil varied task settings and environmental conditions---in just one pass of an optimization algorithm. To this end, in Section II, we first explain the idea of different tasks and environments, and formally define the Set of Sets in this context. In Section III, we then present \emph{neuroevolutionary multitasking} [22--24] as a first of its kind engine for creating the Set of Sets practically and near-optimally. In Section IV, we provide details of our experimental study. In Section V, we report the experimental results together with a summary of additional insights gleaned. Finally, in Section VI we discuss potential directions for future work and conclude the paper.

\section{Introducing the Set of Sets}

Our problem setting revolves around the existence of (a) multiple tasks and (b) multiple environments. It is necessary then to describe each of those components in detail, and what specializing for them entails. We then formally introduce the framework of JATs, MOTs, and MATs. Finally, we tie all of these elements together to formalize the Set of Sets concept. 

\subsection{Multiple Task Settings}
Assimilating fundamental knowledge from multiple tasks can improve predictive performance and model generalization. Hence, after training on a broad spectrum of tasks, JATs become particularly suitable for few-shot and transfer learning [25], making them powerful foundation models [7] which support ML work and research. However, finite-sized JATs still encounter inter-task interferences (eg. conflicting gradients [12]) which could obstruct performance on individual specializations [11].
\begin{figure}[!htbp]
  \includegraphics[width=0.42\textwidth]{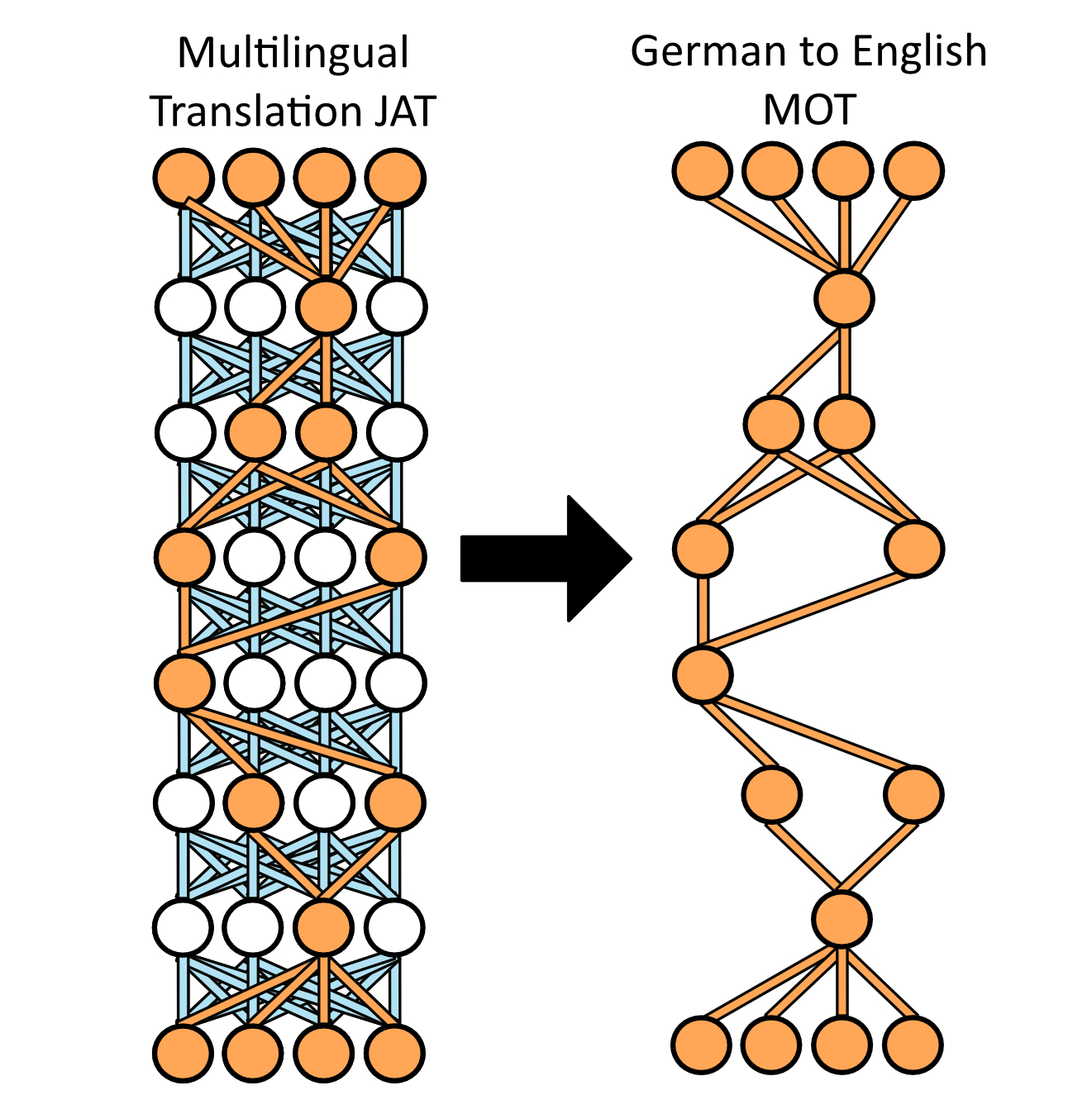}
  \caption{Isolating a compact, task-specialized subnetwork from the large JAT results in a model which could be better at the task and more suitable for deployment outside the cloud. Highlighted connections are those that are retained in the subnetwork.}
  \label{fig:multi-task}
\end{figure}
Moreover, inference costs with JATs could turn out to be a bottleneck, exceeding what is actually needed for the specific objectives and intentions that an end-user may have at any given time. Consider an English-speaking tourist in Germany. Suppose a multilingual translation model has been trained with many northern European languages; these possess high lexical similarity due to common ancestry, resulting in improved translation quality after training. However, the tourist only requires German-to-English translation, not Czech, Swedish or Danish. If German-to-English competence is prioritized above all, then the relevant faculty could be extracted in relative isolation (see illustration in Fig. 2). The main benefit is improved performance and greater efficiency after fine-tuning, assisted by a strong initialization stemming from the model's existing training. 
With that background, we delineate one notion of tasks in what follows.

Consider $K$ supervised machine learning problems, each associated with a labeled dataset $\mathcal{D}^i=(x^{i}_n, y^{i}_n)^{N_i}_{n=1},\; i=1,2,3, ..., K$. It is assumed that all inputs (i.e., $x^i$'s $\forall i$) are embedded in a common vector representation space $\mathcal{X}$, within which the marginal probability distribution $P(x^i)$ may vary across datasets. Here $\mathcal{D}^1$ may pertain to a German-to-English translation, $\mathcal{D}^2$ may pertain to Czech-to-English translation, and so on. A task is thus associated with training an ML model that takes a set of parameters and input data, and outputs a prediction $\hat{y}$ such that $\hat{y} = F(x; \Theta)$. Hereinafter, the predictive function $F$ is assumed to take the form of an arbitrary DNN parameterized by $\Theta$; this could either be a convolutional neural network, an attention-based model, or other neural architecture. The goal is for $F$ to accurately capture the underlying conditional probability distribution in output space $\mathcal{Y}$, i.e., we want $F(x; \Theta) \approx P(y|x)$.  


In order to optimize $\Theta$, tasks are typically assigned a loss  function $\mathcal{L}^{i}$ that encapsulates not only the error between predictions ($\hat{y}$) and ground truth labels ($y^i$)---e.g., cross-entropy loss [26] in the case of classification, focal loss [27] in the case of image segmentation, etc.---but also user-supplied priors and preferences (which could themselves change in time). Thus, we define a task as consisting of the following 3-tuple:    
\begin{equation}
    T_i \triangleq \{\mathcal{D}^i, F(x^{i}; \Theta), \mathcal{L}^i(\hat{y}, y^i)\}.
\end{equation}
Specializing for the $i^{th}$ task setting therefore entails specializing to a single set of these components, by minimizing the loss function $\mathcal{L}^{i}$ over dataset $\mathcal{D}^i$. 

\subsection{Multiple Environmental Conditions}

As more tasks are introduced, performance accuracy is not the only cost incurred. Large-scale models of increasing expressive capacity are required to achieve good average performance across combinations of data distributions. The increasing scale of the best-performing ImageNet models is testament to this phenomenon [28]. Returning to the example of our English-speaking tourist in Germany, the original multilingual model entails significantly greater computational and memory demands compared to a compact monolingual translator. A model this large would typically be housed on a cloud computing cluster, and would depend on real-time connections to provide on-demand service [29]. 

If our hypothetical tourist is expecting unreliable connections, and is operating a typical smartphone dependent on mobile data, then long transmission latencies, connection instability, and high inference times become challenging constraints and factors of concern [30]. All these concerns can be overcome or alleviated by direct deployment of models to mobile devices, as illustrated in Fig. 3. However, directly deploying large-scale cloud-based models on the edge is difficult and often infeasible [16, 17]. This is primarily due to the existence of a very large number of parameters, incurring considerable computational cost and occupying sizeable memory bandwidth. This limitation is acknowledged by growing interest in DNN architectures oriented towards deployment on resource-constrained hardware [31, 32].

Let the structure of a generic DNN, denoted as $\mathcal{G}$, comprise of $L$ layers: $\{\mathcal{G}_1, \mathcal{G}_2, \mathcal{G}_3, ...,\mathcal{G}_L\}$. Each layer $\mathcal{G}_l$ is associated with a parameter vector $\theta_l$ of dimensionality $\rho_l$, where $l$ is the layer index. Since the computing and memory cost of DNNs is strongly dependent on model size [33, 34], one may choose to represent a smaller subnetwork of the full DNN by a binary vector mask $\mathcal{B} = [b^1_1, b^2_1, b^3_1, ..., b^{\rho_1}_{1}, b^{1}_{2}, b^{2}_{2}, b^{3}_{2}, ..., b^{\rho_L}_{L}]$, such that $b^k_{l} \in \{0, 1\}$. Here, $b^k_l = 1$ implies that the $k^{th}$ parameter of the $l^{th}$ layer, i.e., $\theta^k_l$, is retained in the subnetwork, whereas $b^k_l = 0$ represents that the parameter is pruned away (as shown in Fig. 2). This operation is denoted hereafter as $\mathcal{B} \odot \mathcal{G}$.

\begin{figure}[!htbp]
  \includegraphics[width=\linewidth]{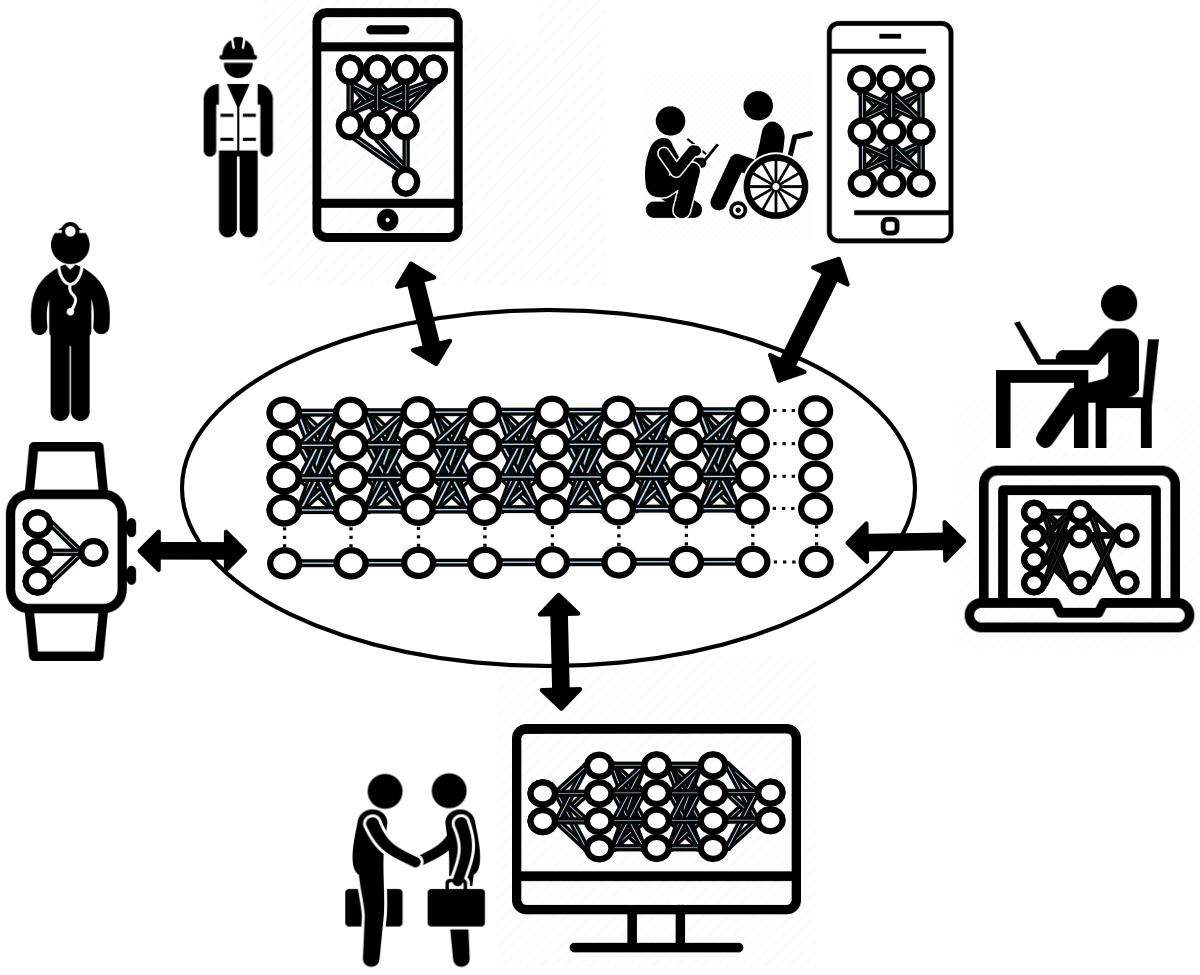}
  \caption{A variety of resource-constrained environmental conditions encountered in practice. Each device demands a particular capability of the original model, but can only support a more restricted model. A large-scale JAT may be infeasible for many such environments.}
  \label{fig:multi-task}
\end{figure}

The $L_1$-norm $||\mathcal{B}||_1$ serves as an approximate representation of the size of a pruned model in terms of the number of parameters it contains. Optimization processes minimizing this norm can perform DNN model compression effectively [35, 36]. This offers a relatively straightforward means of specializing to resource-constrained environments. 

\subsection{Defining MOTs, JATs, and MATs}

Having described the notions of tasks and environments, we may now define the meaning of the terms MOT, JAT, and MAT, in the context of the optimization problem formulation each is associated with.  First, the creation of MOT$_i$ specialized for the $i^{th}$ task can be expressed as follows: 
\begin{equation}
    \begin{gathered}
        \min_{\Theta^i} \frac{1}{N_i}\sum^{N_i}_{n=1} \mathcal{L}^{i}(F(x^{i}_n; \Theta^i), y^{i}_n), \\
        y^{i} \in \mathcal{Y}^{i}, \\
    \end{gathered}
\end{equation}
where $\Theta^{i}$ is the set of model parameters tuned for task $T_i$, and $\mathcal{Y}^i$ is the output label space.

By contrast, a JAT deals with minimizing the average loss over a set of $K$ distinct tasks. As such, it contains a subset of parameters $\Theta^{sh}$ that is strictly shared by all  tasks, in addition to possessing $K$ sets of task-specialized parameters $\Theta^{i}$. The JAT is then formulated as:
\begin{equation}
    \begin{gathered}
  \min_{\Theta^{sh},\; \Theta^{1} \dots \Theta^{K}} \frac{1}{K}\sum^{K}_{i=1} \frac{1}{N_i}\sum^{N_i}_{n=1} \mathcal{L}^{i}(F(x^{i}_n; \Theta^{sh}, \Theta^{i}), y^{i}_n). \\
    \end{gathered}
\end{equation}

The averaging of the loss function over a task set is what gives rise to inter-task interactions. While the resultant inductive transfer can sometimes improve generalization by using information from related tasks, the interference may often hamper individual task performance under finite information encoding capacity of JATs in practice [12]. 

Theoretically, an MAT possesses sufficient capacity to reduce the need for parameter sharing between tasks. Thus, we posit that the MAT could assume at least two different forms. The first is a singular model with sufficient capacity to allocate a dedicated internal substructure of its architecture to each task \emph{when needed}. The second form, that of the Set of Sets, is composed of a plurality of compact MOTs. The MAT could then be regarded as the assimilation of all specialized models, denoted as $\Theta_{MAT} = \cup^{K}_{i=1} \Theta^{i*}$, obtained for all tasks:

\begin{equation}
    \begin{gathered}
    \forall \: T_i, i \in \{1, 2, 3, ..., K\},\\
    \Theta^{i*} = \argmin  \frac{1}{N_i}\sum^{N_i}_{n=1} \mathcal{L}^{i}(F(x^{i}_n; \Theta^{i}), y^{i}_n).\\ 
    \end{gathered}
\end{equation}
This formulation frames a set of MOTs as functionally approximating an MAT. 

\subsection{Formalizing the Set of Sets}

The Set of Sets is a collective of models which approximates an MAT while simultaneously catering to multiple resource-constrained environments. Ideally, it offers a Pareto-optimal model specialized for every combination of task setting and environmental condition. It is thus best to define the Set of Sets, denoted $\mathcal{S}$, in terms of Pareto-optimality [37]. To that end, a model $\Theta^{\alpha}$ is said to be a member of the Set of Sets if there exists some task $T_i$ for which there is no other model $\Theta^{\beta}$ that \emph{dominates} $\Theta^{\alpha}$ in terms of size and performance accuracy. This leads to the following formal definition of \emph{membership in $\mathcal{S}$} in the particular context of models produced by isolating compact subnetworks from a generic DNN $\mathcal{G}$ (i.e., from a large-scale JAT).

\textit{\textbf{Definition 1}} \textit{(Membership in the Set of Sets)}: Let model $\Theta'$ be produced from $\mathcal{G}$ by the operation $\mathcal{B}' \odot \mathcal{G}$. Then, $\Theta^{\alpha} \in \mathcal{S}$ if and only if $\exists \: T_i$ for which $\nexists \: \Theta^{\beta}$ such that $||\mathcal{B}^{\alpha}||_1 \leq ||\mathcal{B}^{\beta}||_1 \land \mathcal{L}^{i}(F(x^{i},\Theta^{\alpha}) \leq \mathcal{L}^{i}(F(x^{i},\Theta^{\beta})$ and $||\mathcal{B}^{\alpha}||_1 < ||\mathcal{B}^{\beta}||_1 \lor \mathcal{L}^{i}(F(x^{i},\Theta^{\alpha}) < \mathcal{L}^{i}(F(x^{i},\Theta^{\beta})$.


Producing such a plurality of $\Theta^{\alpha}$'s is clearly a non-trivial challenge that goes beyond conventional multi-objective optimization problem formulations. The most straightforward solution is to train specialized models individually, from scratch. However, this would incur considerable expense in engineer-labor and energy [20], due to the sheer diversity of possible devices, each with unique resource budgets and requirements [38--40]. This is compounded by each device potentially needing to fulfil a range of task settings. Clearly, a more general, resource-efficient approach is desirable.

\section{Learning the Set of Sets from JATs}
In this section, we offer a means to arrive at the set of ML model sets tractably with just one run of an evolutionary multitasking algorithm.

\subsection{Multi-objective, Multi-task Optimization Formulation}
When trained from scratch for a single task, an MOT does not benefit from a broad base of generalizable knowledge as a foundation model exposed to multifaceted data would. Compressing JATs into MOTs avoids wastage and makes full use of the JAT as a foundation model [7]. Taking these observations and \emph{Definition 1} as the basis of our approach, we consider a formulation that could lead to efficient generation of the Set of Sets by isolating compact subnetworks from a pre-specified JAT $\mathcal{G}$.

Our formulation leverages the dual concepts of multiple tasks and multiple objectives. Referring to Section II-A, we have a series of $K$ tasks, and are interested in minimizing the associated loss functions. This allows us to specialize for multiple task settings concurrently. However, performance is not our sole objective. Referring to Section II-B, we are also interested in optimizing for multiple environmental conditions within the context of each task. Hence, we jointly minimize the $L_1$-norm $||\mathcal{B}||_1$ and the loss function $\mathcal{L}$. This necessitates the acceptance of trade-offs between performance and size, since larger models would generally be expected to perform better but be applicable to only a small range of high-resourced environments. Thus, the Set of Sets solves the following bi-objective optimization problem given a series of tasks: 
\begin{equation}
\begin{gathered}
    \forall T_i,\; i \in \{1, 2, 3, ..., K\}, \\
    \text{minimize}_{\mathcal{B}^{i}}\;\left(||\mathcal{B}^{i}||_1, \frac{1}{N_i}\sum^{N_i}_{n=1} \mathcal{L}^{i}(F(x^{i}_n; \mathcal{B}^i \odot \mathcal{G}), y^{i}_n)\right).\\
\end{gathered}
\end{equation}
This results in sets $PS^i$ of Pareto-optimal models $\Theta^{i*}$ for all  $T_i$, which provide optimized trade-offs of both objective functions. The Set of Sets $\mathcal{S}$ is then the union of all $PS^i$'s:
\begin{equation}
\begin{gathered}
    \Theta^{i*} = \mathcal{B}^{i*} \odot \mathcal{G},\\
    PS^i = \{\Theta^{i*}_1, \Theta^{i*}_2, \Theta^{i*}_3, ...\},\\
    \mathcal{S} = \bigcup^{K}_{i=1} PS^i.
\end{gathered}
\end{equation}
Building a Set of Sets is thus framed as a multi-objective, multi-task optimization problem. By this definition, the Set of Sets is a collective of compressed and specialized DNN models, which offers Pareto-optimal options for a range of environmental conditions and tasks.



\subsection{Other Practical Considerations}
\subsubsection{Mask Dimensionality}
Given the large number of parameters in a modern JAT, the dimensionality of binary mask $\mathcal{B}$ could become a bottleneck---as assigning a binary variable to every parameter can quickly make optimization intractable. One way to overcome this challenge is to reduce the effective dimensionality $\rho_l$ of any layer $l$ by a simple parameter grouping mechanism. One such implementation with masks placed over entire convolutional filters was proposed in [35]. Consider ResNet-18 with its 11M parameters. When each binary variable $b^k_l$ corresponds to an individual parameter, dimensionality is 11,173,962. Alternatively, if each $b^k_l$ groups together approximately half the parameters in layer $l$, then the overall dimensionality of the binary mask is vastly reduced to 36. Note that a group is either wholly retained, if $b^k_l =1$, or wholly pruned away, if $b^k_l =0$. This mask granularity must be carefully selected in a problem dependent manner. 

\begin{algorithm}[h]
\scriptsize
\caption{Pseudocode for Evolving the Set of Sets}
\label{Template}
\textbf{Input}: JAT $\mathcal{G}$, $K$ training datasets $[\mathcal{D}^1, \mathcal{D}^2, \mathcal{D}^3, ..., \mathcal{D}^K]$, population size $\Gamma$, total number of generations $\Omega$: \\
\textbf{Output}: Set of Sets $\mathcal{S}$\\
Randomly generate a population of $\Gamma$ candidate solutions for each of $K$ tasks\\

\For{\text{each population $\mathcal{P}_i,\; i \in \{1, 2, 3, ..., K\}$}}{
\For{\text{each candidate $\mathcal{B}^i_j,\; j \in \{1, 2, 3, ..., \Gamma\}$}}{
Evaluate $\mathcal{B}^i_j$ in terms of $\phi^i_1 = ||\mathcal{B}^i_j||_1$ and $\phi^i_2 = \frac{1}{N_i}\sum^{N_i}_{n=1} \mathcal{L}^{i}(F(x^{i}_n; \mathcal{B}^i \odot \mathcal{G}), y^{i}_n)$ using dataset $\mathcal{D}^i$\\
}
}
$CurrentGen = 0$\\

\While{\text{CurrentGen} $<$ \text{$\Omega$}}{

    Select parent population $\mathcal{P}_i'$ from $\mathcal{P}_i$, $\forall i$ \\ 
    
    
    Assign crossover probability $p_{kl} \in [0, 1]$ between tasks pairs $k$, $l$ \\
    Generate $K$ empty sets of offspring $\mathcal{O}_i$\\

  \For{\text{$j$ in $\{1, 2, 3, ..., \frac{\Gamma \cdot K}{2} \}$}}{
        Randomly select two candidates $\mathcal{B}^k_{\alpha}, \; \mathcal{B}^l_{\beta}$ from $\bigcup^{K}_{i=1} \mathcal{P}_i'$ \\
        Uniformly sample a number $rand \in [0,1]$ \\
        \If{$k == l$}{
            Crossover and mutate $\mathcal{B}^k_{\alpha}, \; \mathcal{B}^l_{\beta}$ to produce two offspring\\
            Evaluate both offspring on $\phi^k_1$ and $\phi^k_2$, and append to $\mathcal{O}_k$ \\ 
        }
        \ElseIf{$k \neq l$ and $rand < p_{kl}$}{
            Crossover and mutate $\mathcal{B}^k_{\alpha}, \; \mathcal{B}^l_{\beta}$ to produce two offspring\\
            Randomly evaluate one offspring on $\phi^k_1$ and $\phi^k_2$, and append to $\mathcal{O}_k$\\
            Evaluate the other on $\phi^l_1$ and $\phi^l_2$, and append to $\mathcal{O}_l$\\
        }
        \Else{
            Randomly select two additional parents $\mathcal{B}^k_{\alpha + 1}$, $\mathcal{B}^l_{\beta + 1}$ from $\mathcal{P}_k$ and $\mathcal{P}_l$, respectively\\
            Crossover and mutate $\mathcal{B}^k_{\alpha}$ with $\mathcal{B}^k_{\alpha + 1}$, evaluate offspring on $\phi^k_1$ and $\phi^k_2$, and append to $\mathcal{O}_k$ \\ 
            Crossover and mutate $\mathcal{B}^l_{\beta}$ with $\mathcal{B}^l_{\beta + 1}$, evaluate offspring on $\phi^l_1$ and $\phi^l_2$, and append to $\mathcal{O}_l$. \\ 
        }
    }
     
     \For{$i$ in $\{1, 2, 3, ..., K\}$}{
     Rank members of the population $\mathcal{O}_i \cup \mathcal{P}_i$\\
     Select $\Gamma$ fittest members from $\mathcal{O}_i \cup \mathcal{P}_i$ to form next $\mathcal{P}_i$ \\
     }
  $\text{$CurrentGen$} = \text{$CurrentGen$} + 1$ ~\\
}

\For{$i$ in $\{1, 2, 3, ..., K\}$}{
     Acquire set of approximated Pareto solutions in $\mathcal{P}_i$ \\
     Fine-tune acquired models by gradient descent on $\phi^i_2$ to get $PS^i$. \\ 
     }

Return $\bigcup^{K}_{i=1} PS^i$
\end{algorithm}

\subsubsection{Layer Compatibility}
It is important to ensure that no layer of the JAT is removed entirely by $\mathcal{B}$, since doing so can collapse performance by cutting off gradient flow. Hence, it is necessary to make sure that a certain number of binary variables $b^k_l$'s remain active in each layer. This defines a hard limit on the extent of compression. Additionally, it is necessary to ensure that redundancies are avoided. For example, it is plausible that all of the input parameters to a node are pruned away, rendering all of that node's output parameters redundant. Constraints and checks to avoid such scenarios or prune away affected parameters in subsequent layers should be implemented.  

\subsection{One-Pass Neuroevolutionary Multitasking}

Multi-objective optimization problems are known to be efficiently handled using Evolutionary Algorithms (EAs) [41, 42]. EAs are crafted to generate and update a diverse population of solutions. When multiple objectives are considered, the population can be structured to approximate Pareto-optimal trade-offs between the objectives [43, 44]; essentially creating a set of solutions that fulfil different environmental conditions or user preferences. However, conventional single- and multi-objective EAs are typically oriented towards solving just a single target task at a time. This tends to limit the \emph{implicit parallelism} and hence the convergence rate of population-based search [45], as useful information or skills learned for other tasks are not leveraged. Evolutionary multitasking offers a natural means to overcome this limitation [46]. 

\begin{figure*}[!htbp]
  \includegraphics[width=0.96\textwidth]{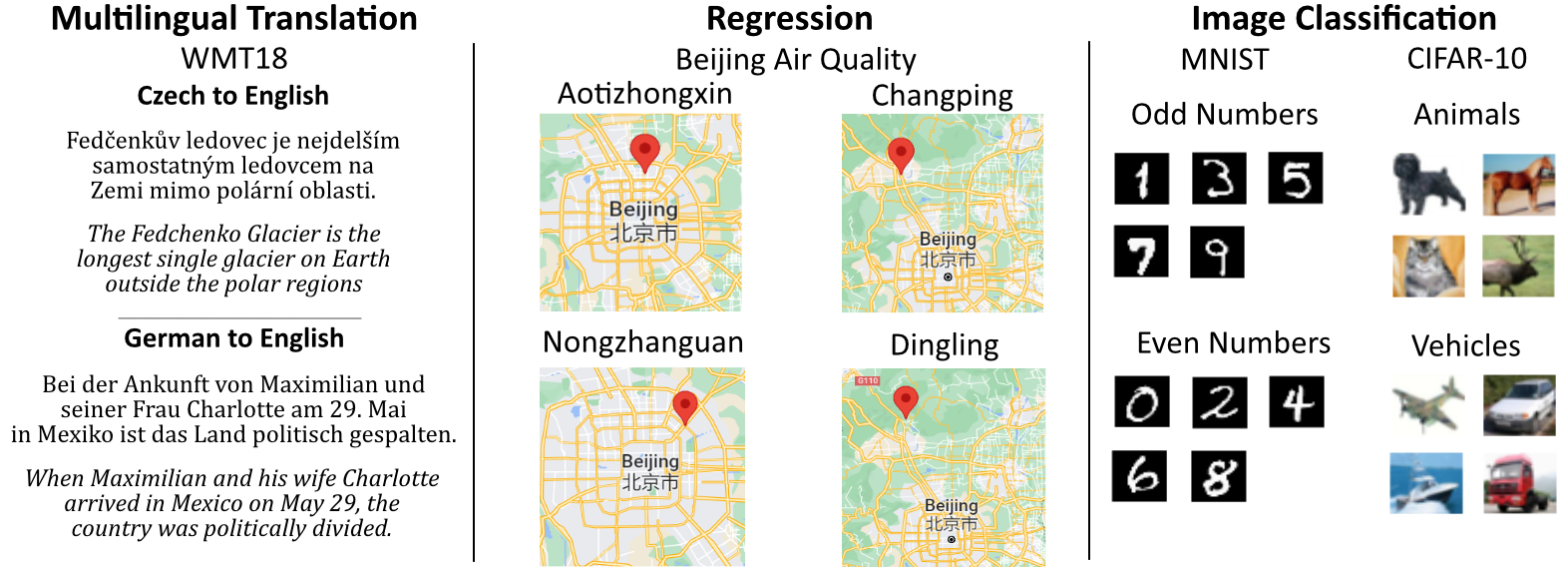}
  \caption{Overview of the tasks used in each experiment. NLP tasks specialize for a single language. Time-series regression tasks specialize for a geographical area. Image classification tasks specialize for a subset of classes.}
  \label{fig:multi-task}
\end{figure*}

Evolutionary multitasking builds on the notion of related tasks sharing reusable building-blocks of knowledge, harnessing the rich body of information gathered while optimizing tasks within the same domain [47, 48]. By transferring useful information across joint optimization processes, it becomes possible to simplify the search, thus speeding up convergence rates to near-optimal solutions [49, 50]. 

Any binary-coded multi-task algorithm can in principle be extended to address all $T_1, T_2, \dots, T_K$ in Eq. (5) in a single run, thus making possible one-pass neuroevolution of the Set of Sets from foundation models. Since the same JAT is assumed to be utilized by all tasks within a domain, a unified solution representation space (i.e., the space of binary masks) is naturally available. Hence, solutions evolved for separate tasks can be easily crossed-over. In producing the Set of Sets, the inter-task crossovers enable the transfer and reuse of subnetworks found to be beneficial across multiple tasks. Consequently, the wastage of computational resources in re-exploring overlapping solution spaces is reduced, enhancing efficiency of the overall optimization cycle [51--53]. 

A generic pseudocode of our approach is given in Algorithm 1. The approach combines the diversity of trade-offs achieved by multi-objective optimization with the improved search generalization of multi-task optimization. In our experimental study, the \emph{multi-objective multifactorial evolutionary algorithm} (MO-MFEA) [23, 54] is employed as an instantiation of an engine for evolutionary multitasking. The MO-MFEA accounts for the multi-objective and multi-task aspects of the formulation in Eq. (5) in a principled manner; it is simple to implement without requiring ad hoc transfer mechanisms [46]. Detailed reviews on alternative algorithmic techniques that can be considered are provided in [55, 56].
\section{Experimental Overview}
In addition to showcasing the effectiveness and efficiency of the problem formulation and approach, the experiments are intended to highlight the generality of the Set of Sets concept across a variety of datasets. The experimental study thus makes use of four datasets, in the domains of multilingual translation, regression, and image classification. A large-scale DNN, representing a JAT, was pre-trained on each dataset in full. Each dataset was then reconfigured into a number of tasks, each of which represents a specialization for categories of data, and is therefore of narrower scope. Specialists are then evolved for all task simultaneously. Fitness is measured using two objective functions: size and performance measure. The final evolved populations are fine-tuned (by gradient descent) on task-specific training data, and then evaluated on held-out test data. All experiments were conducted using an AMD Threadripper 3990X and a single Nvidia 3090 RTX.

\subsection{Multilingual Translation: WMT19}
A unique facet of this work is the application to domains beyond image classification. Following-on on our example of an English-speaking tourist in Germany (see Section II-A), multilingual translation is seen as a natural choice for application of the Set of Sets. Languages with common origins may possess deep fundamental similarity through loan-words and similar lexical structures. Hence, there is a wealth of mutually beneficial information that can be transferred. Specialization is straightforwardly conceptualized---producing master monolingual translators from a large multilingual foundation model. The practical utility of this specialization is evident. 

In keeping with the premise of a JAT being a large, generalist DNN, we make use of Facebook's M2M100-418M model [57] with pretrained weights. The dataset used is the WMT-19 [58] multilingual dataset, focusing on Czech-to-English and German-to-English translation (see Fig. 4). Both of these are acquired from the HuggingFace Transformers and datasets libraries [59]. The pretrained M2M100-418M is trained for both translation tasks for 65 epochs, with a batch size of 1920, until no further loss improvement is observed. The AdamW optimizer [60] is used with a learning rate of 0.0005 and a multiplicative decay of 0.99. 

The dimensionality of the binary mask is 67,584. It is applied solely to attentional layers, and not to vocabulary or position embeddings, or to the prediction head. The model size is approximately 1.8 GB. We then evolve a population of 20 specialist models for each language simultaneously using MO-MFEA, for 120 generations. During evolution, model fitness is assessed using negative log likelihood loss on training data. After termination, we fine-tune the obtained solutions for 200 epochs or until no further improvement is observed,  using the AdaFactor optimizer [61]. After fine-tuning, models are evaluated for BLEU score [62] on the test data.

\begin{figure*}[!htbp]
  \includegraphics[width=0.96\textwidth]{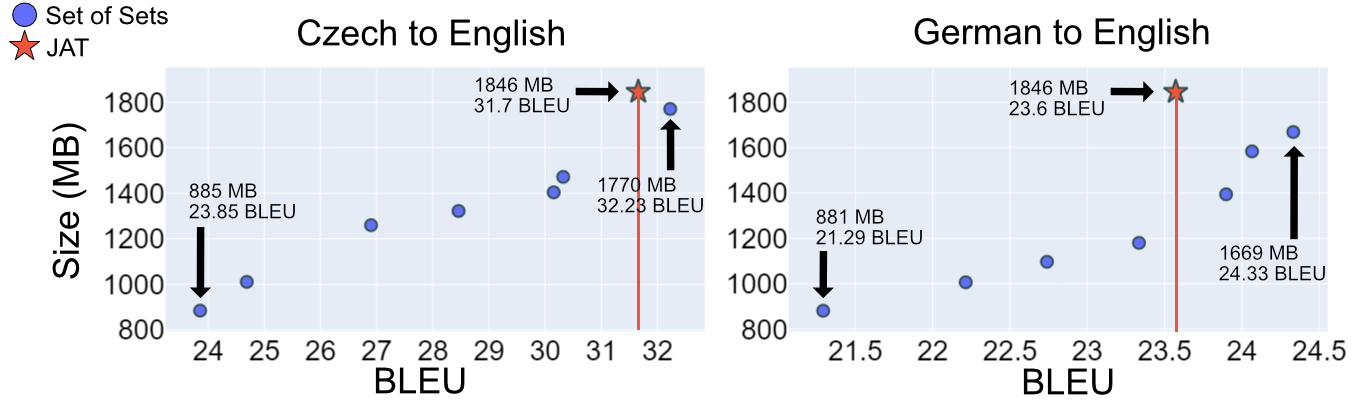}
  \caption{Image in objective space of the evolved Set of Sets for multilingual translation experiments (WMT-19). Each plot displays an evolved population of translators specialized for either Czech or German. Circles denote populations of MOTs, plotted by size and BLEU score for each task. Stars denote the JAT. We achieve MOTs with higher BLEU scores than the JAT, indicating superior performance on the test set. }
  \label{fig:multi-task}
\end{figure*}
\begin{figure*}[!htbp]
  \includegraphics[width=0.96\textwidth]{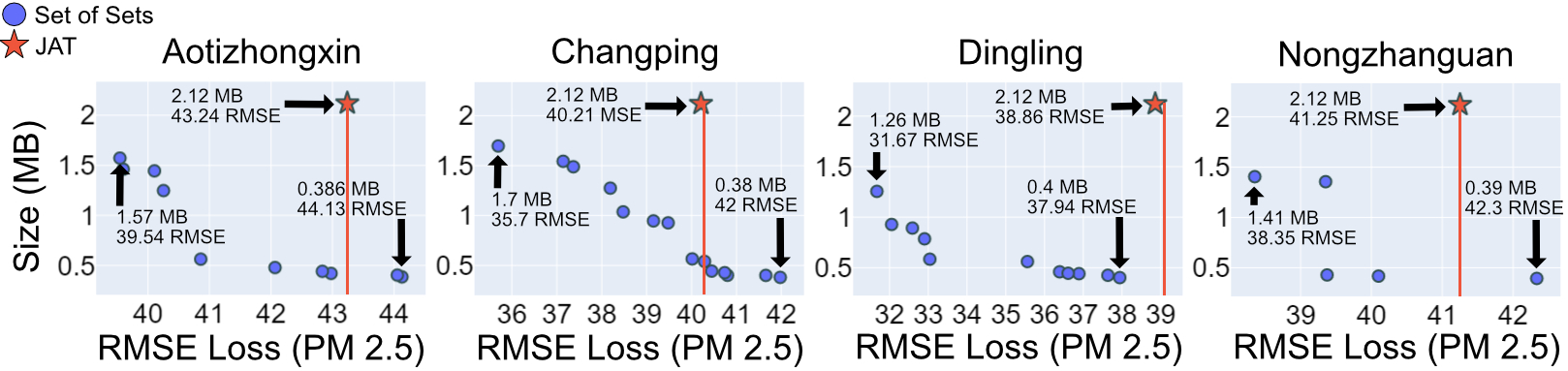}
  \caption{Image in objective space of the evolved Set of Sets for regression (Beijing Air Quality). Each plot displays an evolved population of time-series regression models specialized for air quality prediction in a specific area of Beijing. Circles denote populations of MOTs, plotted by size and RMSE Loss for each task. A large proportion of MOTs achieve lower RMSE Loss than the JAT, potentially making them more useful for predictions in their specific areas.}
  \label{fig:multi-task}
\end{figure*}
\begin{figure*}[!htbp]
  \includegraphics[width=0.96\textwidth]{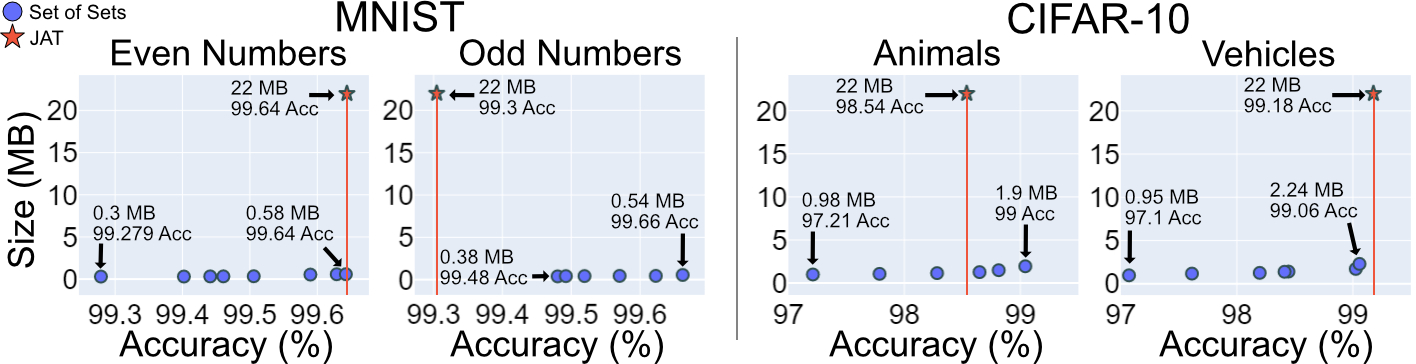}
  \caption{Image in objective space of the evolved Set of Sets acquired for image classification experiments (MNIST and CIFAR-10). Each plot displays an evolved population of classifiers specialized for a subset of classes in their respective datasets. Circles denote populations of MOTs, plotted by size and classification accuracy for each task. In most cases, the presence of MOTs that are significantly more accurate than the JAT is less pronounced than in previous experiments. However, considerable reduction in model size is achieved.}
  \label{fig:multi-task}
\end{figure*}

\subsection{Time-Series Regression: Beijing Air Quality}
Our second domain is time-series regression. In these experiments, we use the Beijing Multi-Site Air-Quality dataset [63] published on UCI. This dataset consists of hourly air-pollutant readings from 12 monitoring sites situated around Beijing. The objective is to predict PM2.5 levels [64]. Specialization is for specific geographic locations (see Fig. 4). Measurements from four monitoring stations, each in a different geographical area, are selected. Datapoints are measured hourly, from March 1st 2013 to February 28th 2017.

A bidirectional LSTM with 2 hidden layers of 128 nodes and one fully-connected output layer is trained on the full dataset [65]. A time window of 10 hours is used. The dimensionality of the binary mask is 4096, and the model is approximately 2.12 MB in size. A population of 60 specialists is evolved corresponding to each area in the MO-MFEA, for 120 generations. After evolution, the obtained solution candidates are fine-tuned for 15 epochs using the AdamW optimizer [60] with a learning rate of 0.001 and a multiplicative decay of 0.99. The root mean squared error (RMSE) loss is used for evolution, fine-tuning, and final evaluation on test data.

\begin{figure*}[!htbp]
  \includegraphics[width=\linewidth]{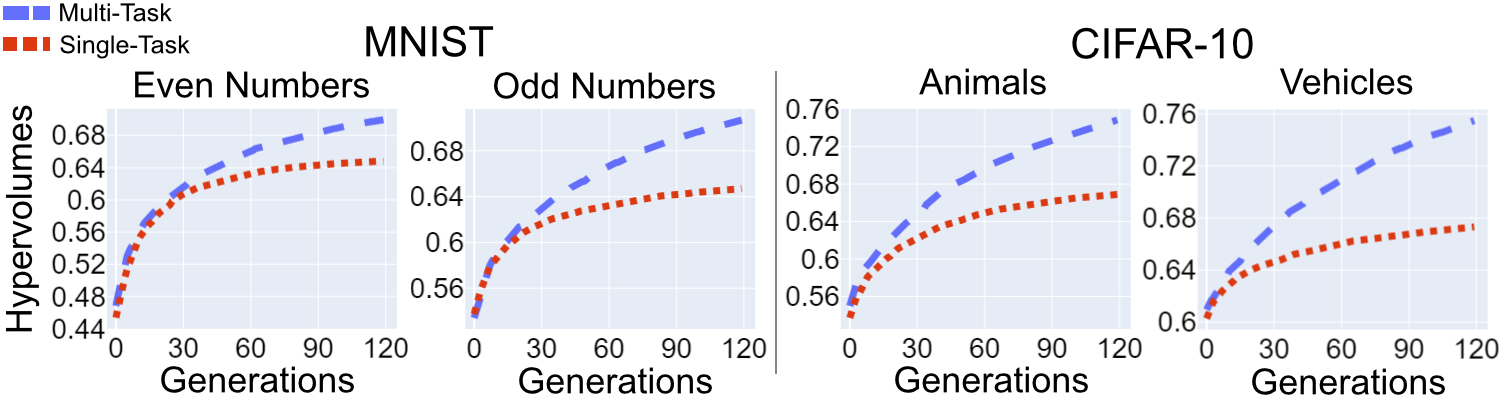}
  \caption{Hypervolume trend comparisons between the multi-task and single-task settings. Multi-tasking results in significantly improved efficiency, requiring less time to converge on a Pareto-optimal population. Multi-tasking can be interpreted as a crucial element in practical usage of the Set of Sets. Note that this experiment made use of automated learning of inter-task similarity measures. Binary mask dimensionality was reduced to 1920, to improve the performance of this automated learning.}
  \label{fig:multi-task}
\end{figure*}

\subsection{Image Classification: MNIST and CIFAR-10}
MNIST [66] and CIFAR-10 [67] are ubiquitous benchmarks in the ML community. As such, we employ them for rigorous comparisons of our neuroevolutionary multitasking approach. As pre-trained ImageNet [68] models constitute an important group of extant foundation models, we use ResNet-18 [69]. This model is approximately 44 MB in size. Additionally, we change the datatype of parameters from 32 bit floats to 16 bit half floats at inference time. This reduces model size to 22 MB. The dimensionality of the binary mask is 18,944. For these datasets, specialized tasks consist of subsets of the labels. MNIST is split into even and odd numbers, while CIFAR-10 is split into animals and vehicles (see Fig. 4). Like the regression experiment, we evolve a population of 60 specialists for each task, for 120 generations. After evolution, the obtained candidates are fine-tuned using the Adam optimizer [70] with a learning rate of 0.0005 and a multiplicative decay of 0.95. MNIST models are fine-tuned for 10 epochs, while CIFAR models are fine-tuned for 25 epochs. The cross-entropy loss is used for evolution and fine-tuning, and the final population is evaluated for classification accuracy on test data. 

\section{Experimental Results}

\subsection{Diverse and Task-Specialized Set of Sets}
Our first experimental results evaluate the Set of Sets against the original JAT. Each evolved population approximates a Pareto front (see Figs. 5-7) showcasing a diversity of trade-offs between model size and performance. In most cases, we observe specialist models which exceed or match the JAT in performance on their assigned tasks. Additionally, we observe significant reductions in model size, while still maintaining acceptable standards of performance, in all tasks considered. Hence, in each domain, the obtained Set of Sets is deemed to collectively approximate an MAT, guaranteeing deployability across a range of resource-constrained environments. 

The most striking examples of performance improvement are observed in the regression tasks (Fig. 6). The best performing regression model, specialized for the extreme North-North-West of Beijing (Dingling), achieves an RMSE reduction of 7.19---a performance improvement of $\sim$18.5\% over the JAT. Note that performance (accuracy) improvement was not witnessed in the CIFAR-10 Vehicles task (Fig. 7). However, the best performing specialist came within 0.12\% accuracy of the JAT. It is likely that further improvement in this case is extremely difficult, given that the underlying JAT appears to have achieved close to 100\% accuracy on this task. 

The clearest examples of successful model compression are seen in the image classification domain, with the smallest specialist for the CIFAR-10 Animals task achieving a size reduction of $\sim$95.5\%. In a less extreme example, the smallest specialist in the Dingling regression task achieves a size reduction of $\sim$82\%. We observe that size reduction is inversely correlated with task difficulty---the NLP tasks are the most difficult and computationally demanding out of those considered, and the smallest specialist in the German-to-English task achieves a size reduction of $\sim$52\% (see Fig. 5). This reveals an interplay between domain, data, and task complexity, as well as model size. We observe that this interplay strongly affects the minimum size that model compression can achieve while still maintaining acceptable levels of performance. 

\begin{figure*}[!htbp]
  \includegraphics[width=\linewidth]{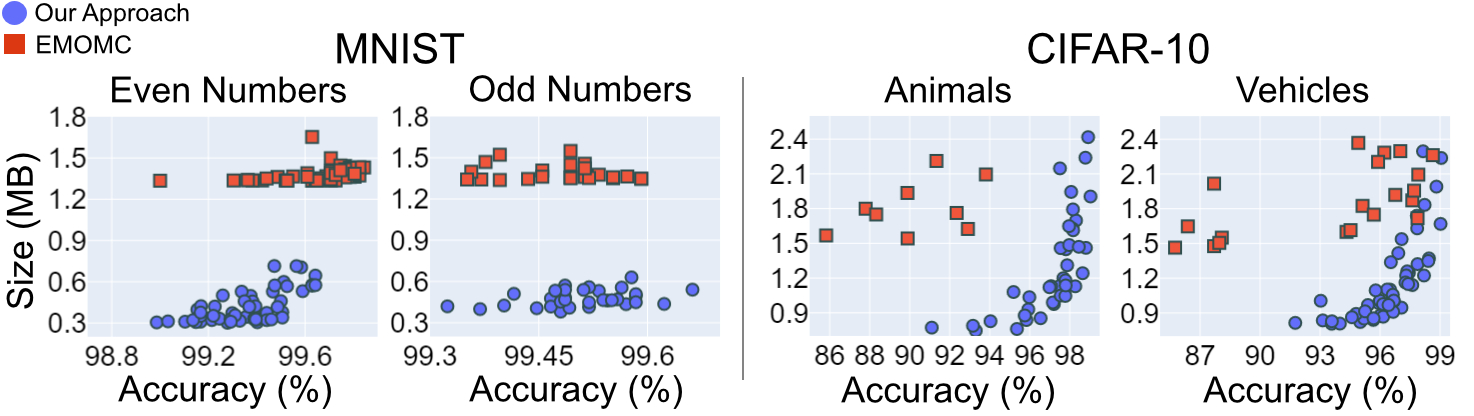}
  \caption{Image in objective space of the entire final population produced by our approach (after fine-tuning), and by [71]. Our approach results in a greater spread of solutions in many cases, with noticeably greater size reductions.}
  \label{fig:multi-task}
\end{figure*}

\subsection{Productivity Gains of One-Pass Learning}

Having demonstrated diversity, we next establish the benefits of neuroevolutionary multitasking for arriving at the Set of Sets in a single optimization run. Conceivably, a comparable Set of Sets could be acquired by repetitively running a single-task multi-objective EA, given sufficient time and resources. As discussed previously, the primary benefit of multi-task optimization lies in transfer of beneficial neural structures between tasks, boosting speed and performance. In other words, given the same compute budget, a neuroevolutionary multitasking algorithm is expected to obtain better solutions than its single-task counterpart. 

For the single-task setting, we disabled the mechanisms of inter-task crossover for information transfer, rendering the MO-MFEA functionally equivalent to the NSGA-II [43]. The hypervolume indicator [72] was employed to assess the quality of populations at each generation. Hypervolume trends were averaged over 10 independent runs of the EAs, providing an indicator of convergence behavior.

In Fig. 8, we observe that hypervolumes increase significantly more rapidly in the multi-task optimization setting compared to the single-task setting. The smallest improvement was seen in the MNIST Even Numbers task, with multitasking achieving a final hypervolume of 0.7 whereas singletasking reached a hypervolume of 0.648 (an $\sim$8\% improvement). The largest improvement was seen in the CIFAR-10 Vehicles task, with MO-MFEA achieving a hypervolume of 0.755 compared to 0.673 in the single-task case (a $\sim$12.2\% improvement). This provides evidence that neuroevolutionary multitasking is able to uncover structural patterns which are commonly beneficial across tasks, and can transfer them across populations to accelerate convergence to positive effect. 

To summarize, it is likely that the single-task approach, given more generations to evolve, will achieve hypervolume scores that are comparable to the MO-MFEA. However, under limited and equal computational budget, the results show that multitasking provides substantial efficiency gains and accelerates convergence to a significant degree.

\subsection{Comparison to the State-of-the-Art}
Here, the competitiveness of our multi-objective, multi-task evolutionary algorithm is established against the state-of-the-art in evolutionary compression. To this end, we make use of the recent EMOMC [71], a single-task, bi-objective evolutionary compression method utilizing both pruning and quantization of DNNs. EMOMC starts with JATs trained on the full MNIST or CIFAR-10 datasets, and subsequently produces a library of pre-pruned models. It then performs evolutionary optimization (selecting pre-pruned models from the library and quantizing their parameters) over the same task-specific datasets as in our approach, hence allowing the generated models to specialize and enabling fair comparison. As with our approach, the EMOMC was run with a population of 60 candidates per task, over 120 generations.

\subsubsection{Results}

As Fig. 9 illustrates, the Set of Sets approach generates a greater spread of possible solutions in many cases. In the MNIST tasks, it is observed that the \emph{average} accuracy of the final population is comparable for both approaches. EMOMC achieves an average accuracy increase of 0.293\% over our approach in the Even Numbers task, while our approach increases the average accuracy by 0.164\% in the Odd Numbers task. On the other hand, in both the CIFAR-10 tasks (Animals and Vehicles), our approach achieves average accuracy increases of 6.83\% and 2.65\%, respectively. 

In addition, the Set of Sets exhibits major improvements in terms of average model size. The most striking examples are in the MNIST Odd Numbers task and the CIFAR-10 Animals task, where our approach achieves an average model size 73.1\% and 38.9\% smaller than EMOMC, respectively. Since EMOMC is a single-task algorithm, these results provide further support to our intuition regarding the benefits of neuroevolutionary multitasking. From a computational standpoint, the Set of Sets approach is indeed more efficient. Each run of the MO-MFEA consumed only about 8 hours, while each run of EMOMC required approximately 23 hours [71].



\subsection{Summarizing Discussions}

The Set of Sets concept was investigated in three diverse and popular domains within deep learning, using a range of datasets, performance metrics, and neural architectures. It was found that on specialized tasks, performance improvements over large-scale models could be achieved simultaneously with compression. In almost all experiments, fine-tuning resulted in specialized models which outperformed the JAT. This indicates that inter-task interference and the trading-off of errors plays a role in constraining model performance. It was also found that evolutionary optimization of the binary masks without fine-tuning is often insufficient to achieve such results, unless the extent of pruning is constrained. Hence, we posit that evolution is able to uncover compact model initializations which respond well to further training. The ensuing fine-tuning then enables the broad and generalizable knowledge contained in the JAT to be efficiently transferred to all MOTs. This makes the Set of Sets approach with neuroevolutionary multitasking practically advantageous, especially when compared to manually training models from scratch for each potential environmental condition and task setting. 


\section{Conclusions}

This paper marks a first study on the concept and in silico evolution of a set of ML model sets from foundation models, pointing to several opportunities for future research. A multi-objective, multi-task problem formulation, and an algorithmic means of arriving at the set in a computationally tractable manner, is offered. The analysis of task-specific neural substructures (their commonalities and distinctions), as they are jointly evolved from a singular JAT, represents an important line of inquiry for the future design of more efficient multi-task algorithms. A comprehensive analysis of the behavior of the Set of Sets, in various task domains (i.e., with changing data distributions, objectives or user intentions), under different environmental conditions, and utilizing a range of possible EAs, is still needed to establish the full extent of its practical value. Finally, the utility of the concept shall be greatly substantiated through implementation in real-world scenarios involving diverse resource-constrained edge devices.

The experiments confirmed that the removal of inter-task interferences from a JAT could indeed prime its compressed counterparts to achieve stronger performance on specialized tasks, with significant reductions in model size. The assimilation of all such compact models gradually approximates what is regarded as a Master of \emph{all} tasks/trades. Additionally, it is shown that the developed techniques can be effectively applied in a diverse range of settings with different neural architectures (spanning transformers for multilingual translation, LSTM for time-series regression, and convolutional networks for image classification), with multitasking greatly improving efficiency in comparison to the associated state-of-the-art.

\bibliographystyle{IEEEtran}

\section{References}

\scriptsize

[1] G.-W. Ng and W. Leung, “Strong artificial intelligence and conscious- ness,” Journal of Artificial Intelligence and Consciousness, vol. 07, pp. 63–72, 03 2020.

[2]  R.  Schwartz,  J.  Dodge,  N.  A.  Smith,  and  O.  Etzioni,  “Green  ai,” Commun. ACM, vol. 63, no. 12, p. 54–63, nov 2020.

[3]  Y.  LeCun,  B.  Boser,  J.  S.  Denker,  D.  Henderson,  R.  E.  Howard, W. Hubbard, and L. D. Jackel, “Backpropagation Applied to Handwritten Zip Code Recognition,” Neural Computation, vol. 1, no. 4, pp. 541–551, 12 1989.

[4] M. Igami, “Artificial intelligence as structural estimation: Deep blue, bonanza, and alphago*,” The Econometrics Journal, vol. 23, 03 2020.

[5] X. Zhai, A. Kolesnikov, N. Houlsby, and L. Beyer, “Scaling vision transformers,” 2021.

[6]  T. Brown, B. Mann, N. Ryder, M. Subbiah, J. D. Kaplan, P. Dhariwal, A. Neelakantan, P. Shyam, G. Sastry, A. Askell, S. Agarwal, A. Herbert- Voss, G. Krueger, T. Henighan, R. Child, A. Ramesh, D. Ziegler, J. Wu,C. Winter, C. Hesse, M. Chen, E. Sigler, M. Litwin, S. Gray, B. Chess, J.  Clark,  C.  Berner,  S.  McCandlish,  A.  Radford,  I.  Sutskever,  and D. Amodei, “Language models are few-shot learners,” in Advances in Neural Information Processing Systems, vol. 33. Curran Associates, Inc., 2020, pp. 1877–1901.

[7] L. Orr, K. Goel, and C. Re´, “Data management opportunities for foundation models,” in 12th Annual Conference on Innovative Data Systems Research, 2021.

[8] R. Bommasani, D. A. Hudson, E. Adeli, R. Altman, S. Arora, S. von Arx, M. S. Bernstein, J. Bohg,  A.  Bosselut,  E.  Brunskill  et  al., “On the opportunities and risks of foundation models,” arXiv preprint arXiv:2108.07258, 2021.

[9] P. Baldi and R. Vershynin, “The capacity of feedforward neural net- works,” Neural networks : the official journal of the International Neural Network Society, vol. 116, pp. 288–311, 2019.

[10] M. Raghu, B. Poole, J. Kleinberg, S. Ganguli, and J. Sohl-Dickstein, “On the expressive power of deep neural networks,” in Proceedings of the 34th International Conference on Machine Learning, ser. Proceedings of Machine Learning Research, vol. 70. PMLR, 06–11 Aug 2017, pp. 2847–2854.

[11] S. Wu, H. R. Zhang, and C. Re´, “Understanding and improving infor- mation transfer in multi-task learning,” in 8th International Conference on Learning Representations, ICLR 2020, Addis Ababa, Ethiopia, April 26-30, 2020.   OpenReview.net, 2020.

[12] T. Yu, S. Kumar, A. Gupta, S. Levine, K. Hausman, and C. Finn, “Gradient surgery for multi-task learning,” in Advances in Neural Information Processing Systems, vol. 33. Curran Associates, Inc., 2020, pp. 5824–5836.

[13] M. G. S. Murshed, C. Murphy, D. Hou, N. Khan, G. Ananthanarayanan, and F. Hussain, “Machine learning at the network edge: A survey,” ACM Comput. Surv., vol. 54, no. 8, oct 2021.

[14] B. Varghese, N. Wang, S. Barbhuiya, P. Kilpatrick, and D. S. Nikolopou- los, “Challenges and opportunities in edge computing,” in 2016 IEEE International Conference on Smart Cloud (SmartCloud), 2016, pp. 20– 26.

[15] H. Ju and L. Liu, “Innovation trend of edge computing technology based on patent perspective,” Wireless Communications and Mobile Computing, vol. 2021, p. 2609700, Jun 2021.

[16] S. Han, H. Mao, and W. J. Dally, “Deep compression: Compressing deep neural network with pruning, trained quantization and huffman coding,” in 4th International Conference on Learning Representations, ICLR 2016, San Juan, Puerto Rico, May 2-4, 2016, Conference Track Proceedings, Y. Bengio and Y. LeCun, Eds., 2016.

[17] K. Bhardwaj, C.-Y. Lin, A. Sartor, and R. Marculescu, “Memory- and communication-aware model compression for distributed deep learning inference on iot,” ACM Trans. Embed. Comput. Syst., vol. 18, no. 5s, oct 2019.

[18]  M. Tegmark, Life 3.0: Being Human in the Age of Artificial Intelligence. Knopf Publishing Group, 2017.

[19] G. F. Marcus, “Deep learning: A critical appraisal,” ArXiv, vol. abs/1801.00631, 2018.

[20] E. Strubell, A. Ganesh, and A. Mccallum, “Energy and policy consider- ations for modern deep learning research,” Proceedings of the AAAI Conference on Artificial Intelligence, vol. 34, pp. 13 693–13 696, 04 2020.

[21] Z. Li, E. Wallace, S. Shen, K. Lin, K. Keutzer, D. Klein, and J. Gonzalez, “Train big, then compress: Rethinking model size for efficient training and inference of transformers,” in Proceedings of the 37th International Conference on Machine Learning, ser. Proceedings of Machine Learning Research, vol. 119.   PMLR, 13–18 Jul 2020, pp. 5958–5968.

[22] A. Gupta, Y.-S. Ong, and L. Feng, “Multifactorial evolution: towards evolutionary multitasking,” IEEE Transactions on Evolutionary Compu- tation, vol. 20, no. 3, pp. 343–357, 2016.

[23] A. Gupta, Y.-S. Ong, L. Feng, and K. C. Tan, “Multiobjective multifac- torial optimization in evolutionary multitasking,” IEEE Transactions on Cybernetics, vol. 47, no. 7, pp. 1652–1665, 2017.

[24] Y.-S. Ong and A. Gupta, “Evolutionary multitasking: a computer science view of cognitive multitasking,” Cognitive Computation, vol. 8, no. 2, pp. 125–142, 2016.

[25] Q. Sun, Y. Liu, T.-S. Chua, and B. Schiele, “Meta-transfer learning for few-shot learning,” in 2019 IEEE/CVF Conference on Computer Vision and Pattern Recognition (CVPR), 2019, pp. 403–412.

[26] Z. Zhang and M. R. Sabuncu, “Generalized cross entropy loss for training deep neural networks with noisy labels,” in Proceedings of the 32nd International Conference on Neural Information Processing Systems, ser. NIPS’18. Red Hook, NY, USA: Curran Associates Inc., 2018, p. 8792–8802.

[27] T. Lin, P. Goyal, R. Girshick, K. He, and P. Dollar, “Focal loss for dense object detection,” IEEE Transactions on Pattern Analysis Machine Intelligence, vol. 42, no. 02, pp. 318–327, feb 2020.

[28]  O.  Russakovsky,  J.  Deng,  H.  Su,  J.  Krause,  S.  Satheesh,  S.  Ma, Z. Huang,  A. Karpathy,  A. Khosla,  M.  Bernstein, A.  C. Berg,  and L. Fei-Fei, “ImageNet Large Scale Visual Recognition Challenge,” International Journal of Computer Vision (IJCV), vol. 115, no. 3, pp. 211–252, 2015.

[29] S. B. Calo, M. Touna, D. C. Verma, and A. Cullen, “Edge computing ar- chitecture for applying ai to iot,” in 2017 IEEE International Conference on Big Data (Big Data), 2017, pp. 3012–3016.

[30]  D. Xu, T. Li, Y. Li, X. Su, S. Tarkoma, T. Jiang, J. Crowcroft, and P. Hui, “Edge intelligence: Architectures, challenges, and applications,” 2020.

[31] M. Sandler, A. Howard, M. Zhu, A. Zhmoginov, and L.-C. Chen, “Mo- bilenetv2: Inverted residuals and linear bottlenecks,” in 2018 IEEE/CVF Conference on Computer Vision and Pattern Recognition, 2018, pp. 4510–4520.

[32]  C. Szegedy, W. Liu, Y. Jia, P. Sermanet, S. Reed, D. Anguelov, D. Erhan, V. Vanhoucke, and A. Rabinovich, “Going deeper with convolutions,” in 2015 IEEE Conference on Computer Vision and Pattern Recognition (CVPR), 2015, pp. 1–9.

[33]  Y.  Tay,  M.  Dehghani,  J.  Rao,  W.  Fedus,  S.  Abnar,  H.  W.  Chung, S. Narang, D. Yogatama, A. Vaswani, and D. Metzler, “Scale efficiently: Insights from pre-training and fine-tuning transformers,” CoRR, vol. abs/2109.10686, 2021.

[34]  J. Kaplan, S. McCandlish, T. Henighan, T. B. Brown, B. Chess, R. Child, S. Gray, A. Radford, J. Wu, and D. Amodei, “Scaling laws for neural language models,” CoRR, vol. abs/2001.08361, 2020.

[35] Y. Wang, C. Xu, J. Qiu, C. Xu, and D. Tao, “Towards evolutionary compression,” in Proceedings of the 24th ACM SIGKDD International Conference on Knowledge Discovery and Data Mining, 2018, pp. 2476– 2485.

[36] Y. Zhou, G. G. Yen, and Z. Yi, “A knee-guided evolutionary algorithm for compressing deep neural networks,” IEEE transactions on cybernet- ics, vol. 51, no. 3, pp. 1626–1638, 2019.

[37] M. Ehrgott, Multicriteria optimization. Springer Science and Business Media, 2005, vol. 491.

[38] H. Cai, C. Gan, T. Wang, Z. Zhang, and S. Han, “Once-for-all: Train one network and specialize it for efficient deployment,” in International Conference on Learning Representations, 2020.

[39] M. Mohammadi, A. Al-Fuqaha, S. Sorour, and M. Guizani, “Deep learning for iot big data and streaming analytics: A survey,” IEEE Communications Surveys Tutorials, vol. 20, no. 4, pp. 2923–2960, 2018.

[40] J. Tang, D. Sun, S. Liu, and J.-L. Gaudiot, “Enabling deep learning on iot devices,” Computer, vol. 50, no. 10, pp. 92–96, 2017.

[41] C. Fan, Y. Li, L. Yi, L. Xiao, X. Qu, and Z. Ai, “Multi-objective lstm ensemble model for household short-term load forecasting,” Memetic Computing, pp. 1–18, 2022.

[42] R. Tanabe and H. Ishibuchi, “A review of evolutionary multimodal multiobjective optimization,” IEEE Transactions on Evolutionary Com- putation, vol. 24, no. 1, pp. 193–200, 2019.

[43] K. Deb, A. Pratap, S. Agarwal, and T. Meyarivan, “A fast and elitist multiobjective genetic algorithm: Nsga-ii,” IEEE transactions on evolu- tionary computation, vol. 6, no. 2, pp. 182–197, 2002.

[44] A. Trivedi, D. Srinivasan, K. Sanyal, and A. Ghosh, “A survey of multiobjective evolutionary algorithms based on decomposition,” IEEE Transactions on Evolutionary Computation, vol. 21, pp. 440–462, 2017.

[45] A. Gupta and Y.-S. Ong, “Back to the roots: Multi-x evolutionary computation,” Cognitive Computation, vol. 11, no. 1, pp. 1–17, 2019.

[46] A. Gupta, L. Zhou, Y.-S. Ong, Z. Chen, and Y. Hou, “Half a dozen real-world applications of evolutionary multitasking, and more,” IEEE Computational Intelligence Magazine, vol. 17, no. 2, pp. 49–66, 2022.

[47] A. Gupta, Y. Ong, and L. Feng, “Insights on transfer optimization: Because experience is the best teacher,” IEEE Transactions on Emerging Topics in Computational Intelligence, vol. PP, pp. 1–14, 11 2017.

[48] A. Gupta and Y.-S. Ong, Memetic computation: the mainspring of knowledge transfer in a data-driven optimization era. Springer, 2018, vol. 21.

[49] L. Bai, W. Lin, A. Gupta, and Y.-S. Ong, “From multitask gradient descent to gradient-free evolutionary multitasking: a proof of faster convergence,” IEEE Transactions on Cybernetics, 2021.

[50] Z. Huang, Z. Chen, and Y. Zhou, “Analysis on the efficiency of multifactorial evolutionary algorithms,” in International Conference on Parallel Problem Solving from Nature.   Springer, 2020, pp. 634–647.

[51] Y.-S. Ong and A. Gupta, “Evolutionary multitasking: A computer science view of cognitive multitasking,” Cognitive Computation, vol. 8, no. 2, pp. 125–142, Apr 2016.

[52] T. Rios, B. van Stein, T. Ba¨ck, B. Sendhoff, and S. Menzel, “Multi- task shape optimization using a 3d point cloud autoencoder as unified representation,” IEEE Transactions on Evolutionary Computation, 2021.

[53] A. D. Martinez, J. Del Ser, E. Osaba, and F. Herrera, “Adaptive multi-factorial evolutionary optimization for multi-task reinforcement learning,” IEEE Transactions on Evolutionary Computation, 2021.

[54] K. K. Bali, A. Gupta, Y.-S. Ong, and P. S. Tan, “Cognizant multitasking in multiobjective multifactorial evolution: Mo-mfea-ii,” IEEE Transac- tions on Cybernetics, 2020.

[55] E. Osaba, A. D. Martinez, and J. Del Ser, “Evolutionary multitask optimization: a methodological overview, challenges and future research directions,” arXiv preprint arXiv:2102.02558, 2021.

[56] T. Wei, S. Wang, J. Zhong, D. Liu, and J. Zhang, “A review on evolutionary multi-task optimization: Trends and challenges,” IEEE Transactions on Evolutionary Computation, 2021.

[57]  A.  Fan,  S.  Bhosale,  H.  Schwenk,  Z.  Ma,  A.  El-Kishky,  S.  Goyal, M. Baines, O. Celebi, G. Wenzek, V. Chaudhary et al., “Beyond english- centric multilingual machine translation,” Journal of Machine Learning Research, vol. 22, no. 107, pp. 1–48, 2021.

[58]  L. Barrault, O. Bojar, M. R. Costa-Jussa, C. Federmann, M. Fishel, Y. Graham, B. Haddow, M. Huck, P. Koehn, S. Malmasi et al., “Findings of the 2019 conference on machine translation (wmt19),” in Proceedings of the Fourth Conference on Machine Translation (Volume 2: Shared Task Papers, Day 1), 2019, pp. 1–61.

[59]  T.  Wolf,  L.  Debut,  V.  Sanh,  J.  Chaumond,  C.  Delangue,  A.  Moi, P. Cistac, T. Rault, R. Louf, M. Funtowicz et al., “Transformers: State- of-the-art natural language processing,” in Proceedings of the 2020 conference on empirical methods in natural language processing: system demonstrations, 2020, pp. 38–45.

[60] I. Loshchilov and F. Hutter, “Decoupled weight decay regularization,” in International Conference on Learning Representations, 2018.

[61] N. Shazeer and M. Stern, “Adafactor: Adaptive learning rates with sublinear memory cost,” in Proceedings of the 35th International Con- ference on Machine Learning, ser. Proceedings of Machine Learning Research, J. Dy and A. Krause, Eds., vol. 80. PMLR, 10–15 Jul 2018, pp. 4596–4604.

[62] K.  Papineni,  S. Roukos,  T.  Ward,  and  W.-J. Zhu,  “Bleu:  a  method for automatic evaluation of machine translation,” in Proceedings of the 40th Annual Meeting of the Association for Computational Linguistics. Philadelphia, Pennsylvania, USA: Association for Computational Lin- guistics, Jul. 2002, pp. 311–318.

[63] S. Zhang, B. Guo, A. Dong, J. He, Z. Xu, and S. X. Chen, “Cautionary tales on air-quality improvement in beijing,” Proceedings of the Royal Society A: Mathematical, Physical and Engineering Sciences, vol. 473, 2017.

[64] A. G. Mengara Mengara, E. Park, J. Jang, and Y. Yoo, “Attention- based distributed deep learning model for air quality forecasting,”  Sustainability, vol. 14, no. 6, 2022.

[65] J. Zhao, F. Deng, Y. Cai, and J. Chen, “Long short-term memory - fully connected (lstm-fc) neural network for pm2.5 concentration prediction,”  Chemosphere, vol. 220, 12 2018.

[66] L. Deng, “The mnist database of handwritten digit images for machine learning research,” IEEE Signal Processing Magazine, vol. 29, no. 6, pp. 141–142, 2012.

[67] A. Krizhevsky, “Learning multiple layers of features from tiny images,” Tech. Rep., 2009.

[68] J. Deng, W. Dong, R. Socher, L.-J. Li, K. Li, and L. Fei-Fei, “Imagenet: A large-scale hierarchical image database,” in 2009 IEEE conference on computer vision and pattern recognition.   Ieee, 2009, pp. 248–255.

[69] K. He, X. Zhang, S. Ren, and J. Sun, “Deep residual learning for image recognition,” in 2016 IEEE Conference on Computer Vision and Pattern Recognition (CVPR), 2016, pp. 770–778.

[70]  D. Kingma and J. Ba, “Adam: A method for stochastic optimization,” ICLR, 12 2015.

[71] Z. Wang, T. Luo, M. Li, J. T. Zhou, R. S. M. Goh, and L. Zhen, “Evo- lutionary multi-objective model compression for deep neural networks,”  IEEE Computational Intelligence Magazine, vol. 16, pp. 10–21, 2021.

[72] A. P. Guerreiro, C. M. Fonseca, and L. Paquete, “The hypervolume in- dicator: Computational problems and algorithms,” ACM Comput. Surv., vol. 54, no. 6, jul 2021.

\end{document}